%% file: main.tex
\documentclass{article}

\usepackage{microtype}
\usepackage{graphicx}
\usepackage{subcaption}
\usepackage{booktabs} % for professional tables

\usepackage{multirow}
\usepackage[table]{xcolor}
\usepackage[most]{tcolorbox}
\usepackage{caption}

\usepackage{hyperref}

\usepackage[preprint]{icml2026}

\usepackage{amsmath}
\usepackage{amssymb}
\usepackage{mathtools}
\usepackage{amsthm}

\usepackage[capitalize,noabbrev]{cleveref}

\theoremstyle{plain}

\theoremstyle{definition}

\theoremstyle{remark}

\usepackage[textsize=tiny]{todonotes}

\icmltitlerunning{MaS-VQA: Mask-and-Select for Knowledge-Based VQA}

\begin{document}

\twocolumn[
\icmltitle{MaS-VQA: A Mask-and-Select Framework for \\ Knowledge-Based Visual Question Answering}

% unless you've provided the [accepted] option to the icml2026


\icmlsetsymbol{equal}{*}

\begin{icmlauthorlist}
\icmlauthor{Xianwei Mao}{zju}
\icmlauthor{Kai Ye}{zju}
\icmlauthor{Sheng Zhou}{zju}
\icmlauthor{Nan Zhang}{ali}
\icmlauthor{Haikuan Huang}{ali}
\icmlauthor{Bin Li}{ali}
\icmlauthor{Jiajun Bu}{zju}
\end{icmlauthorlist}

\icmlaffiliation{zju}{Zhejiang University, Hangzhou, China}
\icmlaffiliation{ali}{Alibaba Group, Hangzhou, China}

\icmlcorrespondingauthor{Sheng Zhou}{zhousheng\_zju@zju.edu.cn}


\vskip 0.3in
]

\printAffiliationsAndNotice{}  % no special notice (required even if empty)

\begin{abstract}
% 最大的问题是当前的模型的问题不够明确，然后你的方法的描述篇幅特别多，这个不是适合在摘要部分写的。

% 写作的基本要求是要说出最最基本的问题，就是你要用外部的知识，但是这个知识因为检索的精度不高，等原因导致你会得到一堆知识知识里面，可是的可能反正最后就是给你的效果来了不好，但是这个其实是低层的，也就是最基础的这个问题
% 但是你要更进一步地让审稿人觉得你的这个技术有难度，或者说觉得这个文章有意思很核心的一点就是，虽然你很容易让别人想到知识会有冗余或者说噪声，但是怎么处理？他有没有什么技术上面的难点和挑战这个所谓的难点和挑战其实是对于你后面模型具体怎么做的一个反过来的说法也是说这是你模型采取的策略，就是因为用别的解决不了，所以你才要设计这个策略。
Knowledge-based Visual Question Answering (KB-VQA) requires models to answer questions by integrating visual information with external knowledge. However, retrieved knowledge is often noisy, partially irrelevant, or misaligned with the visual content, while internal model knowledge is difficult to control and interpret. Naïve aggregation of these sources limits reasoning effectiveness and reduces answer accuracy. To address this, we propose MaS-VQA, a selection-driven framework that tightly couples explicit knowledge filtering with implicit knowledge reasoning. MaS-VQA first retrieves candidate passages and applies a Mask-and-Select mechanism to jointly prune irrelevant image regions and weakly relevant knowledge fragments, producing compact, high-signal multimodal knowledge . This filtered knowledge then guides the activation of internal knowledge in a constrained semantic space, enabling complementary co-modeling of explicit and implicit knowledge for robust answer prediction. Experiments on Encyclopedic-VQA and InfoSeek demonstrate consistent performance gains across multiple MLLM backbones, and ablations verify that the selection mechanism effectively reduces noise and enhances knowledge utilization.
\end{abstract}

\section{Introduction}
\label{sec:intro}
\input{content/1_introduction}

\section{Related Work}
\label{sec:related}
\input{content/2_related}

\section{Method}
\label{sec:method}

\input{content/3_method}

\section{Experiments}
\label{sec:experiment}
\input{content/4_experiment}

\section{Conclusion}
\label{sec:conclusion}
\input{content/5_conclusion}

\bibliography{main_paper}
\bibliographystyle{icml2026}

\clearpage
\appendix
\label{sec:appendix}
\input{content/6_appendix}
\clearpage

\end{document}

%% file: content/1_introduction.tex
Visual Question Answering (VQA) aims to produce an answer by jointly modeling an image and its accompanying question\cite{vqa, vqasurvey}. While many VQA systems have achieved strong performance on questions that can be resolved from the image and the question itself\cite{vqasystem1, qwen3vl}, many real-world questions require additional background knowledge, such as commonsense reasoning, encyclopedic facts, or domain-specific information \cite{evqa, infoseek, survey1}. In such cases, relying solely on the image-question pair is often insufficient, as the required evidence may not be directly observable.
To address this challenge, Knowledge-based Visual Question Answering (KB-VQA) incorporates external knowledge to bridge the semantic and factual gap between visual perception and high-level reasoning. By jointly reasoning over images and knowledge sources, KB-VQA enables more accurate answers to knowledge-intensive questions.

\begin{figure}[!t]
  \centering
  \includegraphics[width=\columnwidth]{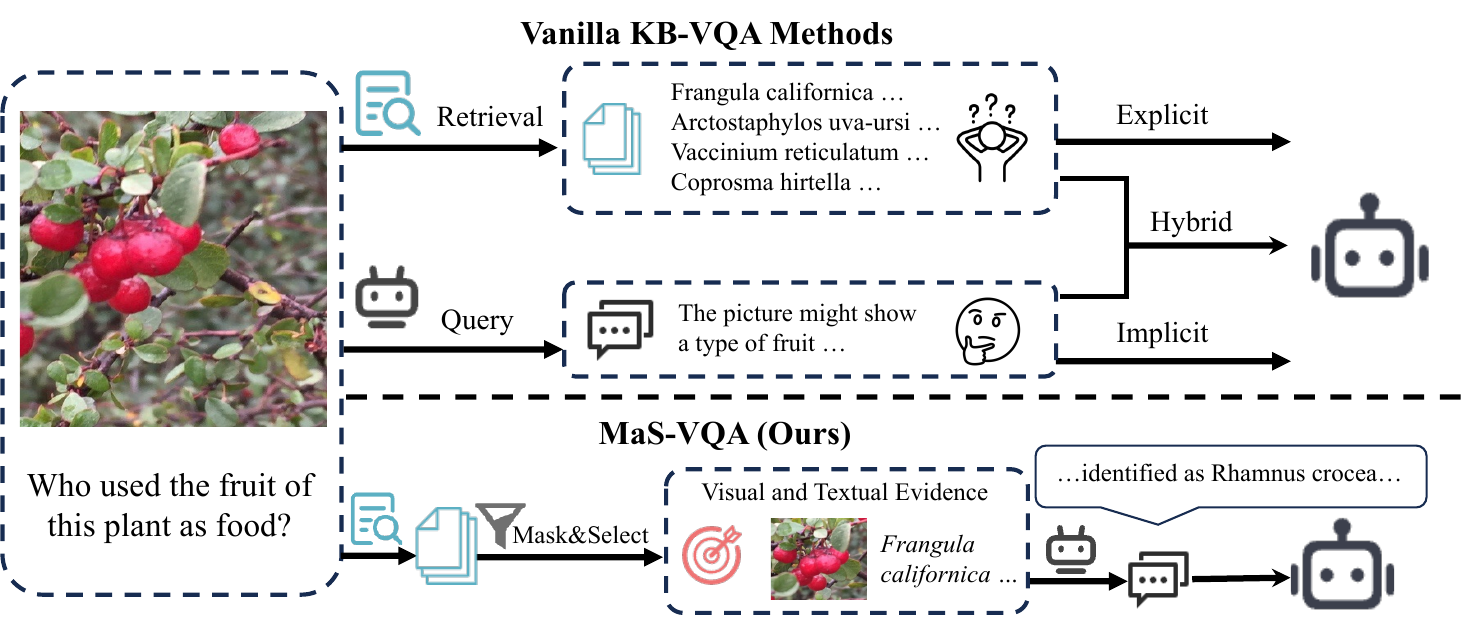}
  \caption{A comparison of vanilla KB-VQA and our proposed method. Compared to standard hybrid methods that separate explicit and implicit knowledge, MaS-VQA integrates their reasoning.}
  \label{fig:introfigure}
\end{figure}

Existing KB-VQA methods can be broadly categorized into three paradigms: explicit, implicit, and hybrid approaches, as illustrated in Figure~\ref{fig:introfigure}. Explicit methods align visual entities or regions with external knowledge sources \cite{vlcbert, echosight}, while implicit methods rely on internal parametric knowledge embedded in large-scale pre-trained models for cross-modal reasoning \cite{pica, prophet}. Hybrid approaches attempt to combine retrieved external knowledge with internal parametric knowledge to leverage complementary strengths \cite{kat, notemr}. Despite differences in implementation, all these paradigms aim to enhance KB-VQA by augmenting visual understanding with external or latent knowledge.

Even with this progress, a central challenge in KB-VQA remains unresolved: how to effectively control which visual knowledge and which retrieved knowledge are jointly used for reasoning under noisy and heterogeneous inputs. 
In practice, both visual and retrieved knowledge are often noisy—e.g., object detectors typically generate overlapping or repetitive region candidates\cite{yolov10}, and knowledge retrievers tend to return partially irrelevant or semantically duplicated fragments due to limited retrieval accuracy and imperfect query formulation\cite{echosight}. Their relevance is often cross-modally coupled: the importance of a visual region may depend on specific textual knowledge, while the usefulness of textual knowledge is conditioned on visual context.
However, many existing methods rely on coarse or single-step filtering strategies that treat visual and textual relevance independently, which limits effective integration between explicit knowledge and implicit parametric reasoning \cite{img2llm, gerea, wwwworkshop}.
These limitations ultimately hinder fine-grained, modality-aware control of cross-modal knowledge exposure, making it difficult to support meaningful interaction between explicit retrieved knowledge and implicit parametric knowledge.

% 前面这段压缩之后，这里可以开一小段介绍challenge，也可以合并到前面
% 因为冗余/噪声是很基础的问题，而且现有方法并不是啥也没干，所以我们不能仅仅停留在喷冗余这个层次，这样相当于否定了现有方法在处理冗余上的贡献，虽然人家可能没有把它当成核心contribution，但是并不是啥也没干
% 因此需要说一下更为深入的难点，也就是说，同样是解决这个冗余/噪声，为什么现有方法不行，这个才是核心
% 所谓的深入的就是你现在设计的这个机制的动机，简单的筛选knowledge怎么就不对了，为什么要设计这么“复杂”的机制来反复mask-and-select，现在最大的问题是看完之后没看出来为什么要mask-and-selct，只知道确实冗余，但是不能让人觉得就应该要用mask-and-select，原因还是缺少中间的“更为深入”的问题，这个问题挖出来了，才能让人同意你的mask-and- select，例如图像，文本，为什么要用不同的机制？两者是否有先后顺序？以及组合起来的难点？
% 上一段的篇幅可以干掉50%，留给这一段更为重要的“challenge”或者“问题”，因为上面那一段的问题比较浅，大家都知道，而且和你的设计没有因果关系，只是说你的方法能部分解决上面的问题，不代表看完上面的问题，就觉得“应该”用你的解法

Building on this insight, we propose \textbf{MaS-VQA}, a selection-based framework that tightly couples explicit knowledge filtering with implicit parametric reasoning, as illustrated in Figure~\ref{fig:introfigure}.
Given an image–question pair, MaS-VQA first retrieves top-$k$ candidate passages via a multimodal retriever. It then performs explicit knowledge processing through a unified \textbf{Mask-and-Select} mechanism that jointly filters visual and textual knowledge: on the visual side, a question-conditioned cross-attention module produces a knowledge-guided attention mask to suppress irrelevant regions; on the textual side, question-conditioned phrase selection preserves key knowledge fragments while masking noisy or weakly relevant content.
Building upon the filtered multimodal knowledge, MaS-VQA further performs implicit knowledge processing conditioned on the refined explicit knowledge, guiding the model to activate and reason over relevant model-internal knowledge within a more constrained semantic space. In this way, explicit and implicit knowledge are co-modeled as complementary sources for answer prediction, yielding more robust KB-VQA under noisy retrieval and complex inputs.

We evaluate MaS-VQA on two challenging KB-VQA benchmarks, Encyclopedic-VQA~\cite{evqa} and InfoSeek~\cite{infoseek}. Experimental results show that MaS-VQA consistently improves KB-VQA performance across different MLLM backbones, demonstrating strong robustness to noisy retrieval and complex inputs. Further analyses and ablations verify the effectiveness of each component in MaS-VQA, and qualitative visualizations illustrate that our selection mechanism yields more focused visual grounding and more reliable knowledge utilization. Our contributions can be summarized as follows:
\begin{itemize}
\item We propose MaS-VQA, a selection-based framework that tightly couples explicit knowledge filtering with implicit parametric reasoning for KB-VQA.
\item We introduce a unified \textbf{Mask-and-Select} mechanism that performs fine-grained selection over both visual regions and retrieved knowledge, producing compact, high-signal explicit representations to mitigate noise accumulation.
\item We conduct comprehensive experiments on Encyclopedic-VQA and InfoSeek, achieving consistent improvements and providing detailed ablations and qualitative analyses to validate the proposed design.
\end{itemize}

%% file: content/2_related.tex
% The primary distinction between KB-VQA and standard VQA is the explicit incorporation of external knowledge to enable reasoning beyond what is directly observable in the image. Consequently, 
KB-VQA methods fall into three categories depending on how knowledge is acquired and used, namely explicit, implicit, and hybrid. We review each in turn.

\subsection{Explicit Knowledge Methods}
Explicit knowledge methods retrieve facts from external knowledge bases and fuse them into answer prediction.

Early methods form KB queries from visual and textual cues, such as VLC-BERT\cite{vlcbert} which concatenates detected object names with the question. Later work improves retrieval granularity and representations. KAT\cite{kat} matches image patches to knowledge snippets, and RKVQA\cite{rkvqa} jointly embeds global and local regions for retrieval. Separately, REVEAL\cite{reveal} maintains updatable explicit knowledge via a key–value memory.

Explicit knowledge methods offer controllable, verifiable evidence, but their performance hinges on query formulation, retrieval granularity, and multimodal fusion.

\begin{figure*}[t]
  \centering
  \includegraphics[width=\textwidth]{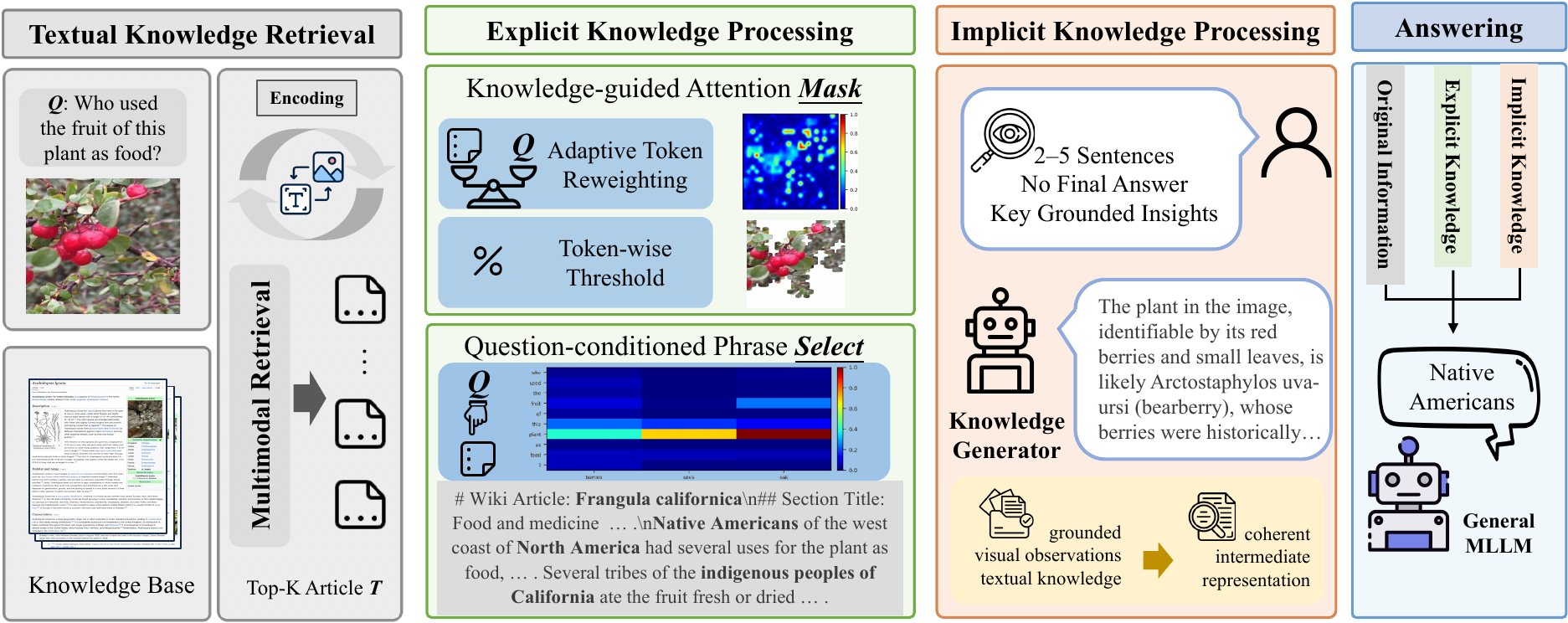}
  \caption{Overview of MaS-VQA. Given an image--question pair, MaS-VQA retrieves top-$k$ passages from an external knowledge base and performs \textbf{Mask-and-Select} explicit knowledge processing, including a knowledge-guided attention mask for filtering irrelevant image regions and question-conditioned phrase selection for pruning noisy text. The filtered multimodal evidence is then used for implicit knowledge processing to elicit complementary model-internal knowledge, and both knowledge sources are co-modeled for final answer prediction.}
  \label{fig:methodfigure}
\end{figure*}

\subsection{Implicit Knowledge Methods} 
Implicit knowledge methods rely on knowledge stored in model parameters, often elicited through prompting large language models.

Several works activate implicit knowledge via captioning, few-shot prompting, and pre-answer. PICa\cite{pica} prompts GPT-3 with image captions and few-shot examples, while ASB\cite{asb} uses the same strategy to elicit implicit knowledge in LLaMA. Beyond direct captions, GeReA\cite{gerea} uses Grad-CAM to guide caption generation for MLLM prompting, Img2LLM\cite{img2llm} augments prompts with related QA pairs, and PromptCap\cite{promptcap} distills GPT-3-style question-conditioned captioning into a lightweight captioner for few-shot reasoning. Complementarily, Prophet\cite{prophet} introduces an upstream lightweight VQA model to produce a pre-answer that, together with few-shot prompting, better unlocks GPT-3’s capabilities, and PCPA\cite{pcpa} adopts a similar pre-answer pipeline for LLaMA.

Recent methods improve knowledge selection and self-learning. ReflectiVA\cite{reflectiva} lets the model decide whether to retrieve external evidence or rely on internal knowledge, while HinD\cite{hind} enhances MLLM self-learning to improve knowledge generation and reasoning.

Implicit methods reduce reliance on external knowledge-base coverage, but they remain prone to hallucination and may lag in factual freshness, which motivates hybrid designs.

\subsection{Hybrid Explicit-implied Knowledge Methods} 
Many recent systems combine explicit retrieval with LLM/MLLM implicit knowledge to support knowledge-intensive reasoning. KAT~\cite{kat} couples patch-level visual retrieval with caption-based prompting for GPT-3 to inject external evidence, while KRISP~\cite{krisp} jointly learns explicit concept-graph reasoning together with implicit answer-space priors to improve robustness. RKVQA~\cite{rkvqa} integrates both global and local retrieval signals and leverages caption prompting to bridge retrieved content and question answering. NoteMR~\cite{notemr} further pairs an explicit retriever with direct MLLM QA, treating the model itself as a complementary knowledge source and refining predictions based on both retrieved notes and implicit model knowledge.

%% file: content/3_method.tex
In this section, we introduce our proposed MaS-VQA framework. In Sec.~\ref{subsec:taskformulation}, we formalize the task and define the answer prediction objective under an external knowledge source.
In Sec.~\ref{subsec:explit}, we describe how to construct the explicit knowledge package $E=\{T,\mathbf{k},M\}$, including retrieving top-$k$ passages, selecting question-relevant keyword phrases, and generating a knowledge-guided attention mask for visual grounding.
In Sec.~\ref{subsec:impliciit}, we introduce implicit knowledge generation, where a frozen MLLM distills a concise grounded paragraph conditioned on $(I,Q,E)$ to complement the retrieved evidence.
Fig.~\ref{fig:methodfigure} illustrates the overall architecture of MaS-VQA.

\subsection{Task Formulation}
\label{subsec:taskformulation}
Given an image $I$ and a question $Q$, KB-VQA aims to predict an answer $\hat{A}\in\mathcal{A}$ by leveraging an external knowledge source $\mathcal{K}$ in addition to visual evidence:
\begin{equation}
\hat{A}=\arg\max_{a\in\mathcal{A}} p(a\mid I,Q,\mathcal{K}).
\end{equation}

As shown in Fig.~\ref{fig:methodfigure}, our framework constructs \textbf{explicit knowledge} from retrieval and \textbf{implicit knowledge} generated by a frozen multimodal large language model (MLLM).
We first retrieve a set of top-$k$ candidate knowledge snippets as textual knowledge:
\begin{equation}
T=\{t_i\}_{i=1}^{k}=f_{\mathrm{ret}}(I,Q;\mathcal{K}),
\end{equation}
where each $t_i$ is a retrieved passage (e.g., a Wikipedia segment).

To facilitate grounding of the retrieved knowledge to the image, we select a keyword set $\mathbf{k}$ from $T$ and derive an attention map (mask) $M$ that highlights question-relevant image regions under the guidance of retrieved text:
\begin{equation}
\mathbf{k}=f_{\mathrm{key}}(T),
\qquad
M=f_{\mathrm{mask}}(I,Q,T).
\end{equation}

For notational convenience, we denote the explicit knowledge package as
\begin{equation}
E=\{T,\mathbf{k},M\}.
\end{equation}
Conditioned on the image, question, and explicit knowledge, we generate an \emph{implicit knowledge} paragraph $U$ (2--5 sentences) that captures concise grounded insights useful for answering the question:
\begin{equation}
U=f_{\mathrm{imp}}(I,Q,E).
\end{equation}

Finally, we query a frozen MLLM $g_{\theta}$ with all evidence and obtain the answer distribution:
\begin{equation}
p(a\mid I,Q,\mathcal{K}) = g_{\theta}(a \mid I,Q,E,U),
\end{equation}
\begin{equation}
\hat{A}=\arg\max_{a\in\mathcal{A}} p(a\mid I,Q,\mathcal{K}).
\end{equation}
Here, $E$ corresponds to explicit knowledge processing (retrieval and grounding), while $U$ represents implicit knowledge distilled under explicit guidance for robust final reasoning.

\subsection{Explicit Knowledge Processing: Text and Image Signals}
\label{subsec:explit}
Given the retrieved knowledge passages $T=\{t_i\}_{i=1}^{k}$ and the question $Q$, we construct two explicit grounding signals: (i) an \emph{image-side} attention mask $M$ that localizes question-relevant visual regions under the guidance of the retrieved text, and (ii) a \emph{text-side} keyword phrase set $\mathbf{k}$ that condenses the retrieved knowledge into a few high-salience spans. In practice, both signals are extracted from a pretrained image-text matching (ITM) encoder.

We tokenize the knowledge and question separately and concatenate them into a single sequence:
\begin{equation}
X=\texttt{[CLS]}\;T\;\texttt{[SEP]}\;Q\;\texttt{[SEP]},
\end{equation}
where we also keep the character-level offset mapping for knowledge tokens (enabled by using a fast tokenizer). Let $\mathcal{I}_K=[k_s,k_e)$ and $\mathcal{I}_Q=[q_s,q_e)$ denote the token index ranges of knowledge and question in $X$, determined by the separator token positions. When the sequence exceeds the maximal text length, we truncate $X$ and correspondingly clip the effective knowledge offsets.

\paragraph{Image-side: knowledge-guided attention mask generation.}
We compute an image patch mask that highlights regions supported by the joint evidence of the knowledge $T$ and question $Q$.
Let the visual encoder produce $P$ patch embeddings ($P=g^2$ for a $g\times g$ patch grid), and let the text encoder attend to visual patches through cross-attention.
At a chosen transformer block $b$, we extract the cross-attention weights
$\mathbf{A}^{(b)}\in\mathbb{R}^{H\times L\times P}$
and their backpropagated sensitivity signals $\nabla \mathbf{A}^{(b)}$ with respect to the positive ITM logit.
We construct token-to-patch relevance by head aggregation and non-negative filtering:
\begin{equation}
\mathbf{R}_{i,p}=\frac{1}{H}\sum_{h=1}^{H}\Big(\mathbf{A}^{(b)}_{h,i,p}\cdot \mathrm{ReLU}(\nabla \mathbf{A}^{(b)}_{h,i,p})\Big),
\end{equation}
where $i\in\{1,\dots,L\}$ indexes text tokens and $p\in\{1,\dots,P\}$ indexes visual patches.
We reshape $\mathbf{R}$ into token-wise $g\times g$ maps.
For numerical stability, each token map is min--max normalized to $[0,1]$ independently.

\paragraph{Adaptive token reweighting.}
Different tokens in $T$ and $Q$ may contribute unequally to localization.
We compute a token strength score by averaging its relevance over patches:
\begin{equation}
s_i=\frac{1}{P}\sum_{p=1}^{P}\mathbf{R}_{i,p}.
\end{equation}
Let $\mathcal{I}_K$ and $\mathcal{I}_Q$ denote the token index sets corresponding to knowledge and question spans, respectively.
We derive intra-group weights via a temperature-controlled softmax, applied separately within $\mathcal{I}_K$ and $\mathcal{I}_Q$:
\begin{equation}
\tilde{w}_i=
\begin{cases}
\mathrm{softmax}\!\left(\frac{s_i}{\tau}\right), & i\in \mathcal{I}_K,\\
\mathrm{softmax}\!\left(\frac{s_i}{\tau}\right), & i\in \mathcal{I}_Q,\\
0, & \text{otherwise}.
\end{cases}
\end{equation}
To balance the overall contributions of knowledge and question tokens, we further compute a group-level adaptive factor:
\begin{align}
\alpha &= \mathrm{softmax}\!\left(\frac{[\bar{s}_K,\;\bar{s}_Q]}{\tau}\right), \\
\bar{s}_K &= \frac{1}{|\mathcal{I}_K|}\sum_{i\in\mathcal{I}_K} s_i, \qquad
\bar{s}_Q = \frac{1}{|\mathcal{I}_Q|}\sum_{i\in\mathcal{I}_Q} s_i .
\end{align}
We set $w_i=\alpha_K\tilde{w}_i$ for $i\in\mathcal{I}_K$ and $w_i=\alpha_Q\tilde{w}_i$ for $i\in\mathcal{I}_Q$, and normalize the weights such that $\sum_{i=1}^{L} w_i=1$.

\paragraph{Token-wise thresholding and patch mask composition.}
We first form weighted token-to-patch scores
\begin{equation}
\hat{\mathbf{R}}_{i,p} = w_i\,\mathbf{R}_{i,p}.
\end{equation}
Instead of thresholding a single aggregated patch score, we perform \emph{token-wise} percentile thresholding.
For each token $i$, we compute a token-specific threshold as the $\rho$-percentile of $\{\hat{\mathbf{R}}_{i,p}\}_{p=1}^{P}$ and obtain a binary token mask over patches:
\begin{equation}
\mathbf{B}_{i,p}=\mathbb{I}\!\left[\hat{\mathbf{R}}_{i,p}>\mathrm{Quantile}\!\left(\hat{\mathbf{R}}_{i,:},\rho\right)\right].
\end{equation}
Finally, we compose the overall patch mask by combining token masks across all tokens using a logical \textsc{OR} operator:
\begin{equation}
M_p = \bigvee_{i=1}^{L}\mathbf{B}_{i,p}, \qquad p=1,\dots,P.
\end{equation}
$M$ is reshaped to a $g\times g$ patch mask, upsampled to the image resolution, and used for visualization by suppressing non-salient regions with a white background.

\paragraph{Text-side: question-conditioned phrase selection from retrieved knowledge.}
We additionally select a compact set of phrases from $T$ to serve as high-precision textual hints. Using the same tokenized sequence $X$, we compute the self-attention probabilities at a selected text self-attention layer:
\begin{equation}
\mathbf{A}\in\mathbb{R}^{H\times L\times L}.
\end{equation}
To quantify how strongly each knowledge token supports the question, we construct a question-to-knowledge interaction matrix by modulating $\mathbf{A}$ with its sensitivity signal and aggregating heads:
\begin{equation}
\mathbf{S}=\frac{1}{H}\sum_{h=1}^{H}\Big(\mathbf{A}_{h}\odot \mathrm{ReLU}(\nabla \mathbf{A}_{h})\Big)\in\mathbb{R}^{L\times L},
\end{equation}
and compute a score for each knowledge token $j\in\mathcal{I}_K$ by averaging over question tokens:
\begin{equation}
s_j=\frac{1}{|\mathcal{I}_Q|}\sum_{i\in\mathcal{I}_Q}\mathbf{S}_{i,j}.
\end{equation}
We select the top-$m$ knowledge tokens with the largest $s_j$, and map these token indices back to character spans in the original knowledge text using the offset mapping. Since selected tokens may be fragmented, we merge overlapping or adjacent spans within a small gap to form readable phrases. The final keyword set $\mathbf{k}$ is the collection of merged spans selected from $T$, and is used as the text-side explicit signal in the downstream prompting stage.

\subsection{Implicit Knowledge Processing}
\label{subsec:impliciit}
Large multimodal language models (MLLMs) pretrained on massive image-text corpora encode substantial world knowledge and reasoning priors in their parameters. However, directly answering KB-VQA questions with a frozen MLLM can be unreliable when the required evidence is implicit, multi-hop, or needs to be grounded to specific image regions. To better exploit the model's internal knowledge while maintaining faithfulness to the given evidence, we distill an \emph{implicit knowledge} paragraph $U$ as an intermediate representation conditioned on the retrieved explicit knowledge and its visual grounding.

\paragraph{Inputs and format.}
Following Fig.~\ref{fig:methodfigure} and the definition in Sec.~\ref{subsec:taskformulation}, we generate the implicit knowledge paragraph $U$ by conditioning a frozen MLLM on the image $I$, the question $Q$, and the explicit knowledge package $E=\{T,\mathbf{k},M\}$.
In practice, we instantiate this step with a structured instruction prompt that includes:
(i) the retrieved passages $T$; (ii) the selected keywords $\mathbf{k}$; and (iii) an attention-map-guided image view induced by $M$ (non-salient regions are suppressed).
The MLLM is instructed to output a concise paragraph (2--5 sentences) that summarizes grounded clues and intermediate conclusions useful for answering $Q$.

\paragraph{Role of implicit knowledge.}
The generated $U$ serves two purposes.
First, it compresses long and potentially noisy retrieved passages into a short, question-focused hypothesis space, improving robustness to retrieval errors.
Second, it integrates grounded visual observations (guided by $M$) with textual knowledge (from $T$ and $\mathbf{k}$), producing a coherent intermediate representation that is easier for the final answering step to consume.

\paragraph{Usage in final inference.}
The implicit knowledge $U$ is treated as supplementary evidence and is appended to the final reasoning prompt together with $(I,Q,E)$ when querying the frozen MLLM for the answer distribution.

%% file: content/4_experiment.tex
\begin{table*}[t]
\caption{Performance on the Encyclopedic-VQA test set and the InfoSeek validation set. Best results are highlighted in bold. $^{\ast}$ denotes methods that require additional training during the reasoning stage. $^{\dagger}$ denotes results that are not directly comparable due to differences in the underlying knowledge bases. $^{\lozenge}$ denotes our reproductions using different MLLM backbones.}
\centering
\small
\setlength{\tabcolsep}{3pt}
\resizebox{\textwidth}{!}{
\begin{tabular}{l c c c c c c c}
\toprule
\multirow{2}{*}{\textbf{Method}} &
\multirow{2}{*}{\textbf{Model}} &
\multirow{2}{*}{\textbf{Retrieval Mode}} &
\multicolumn{2}{c}{\textbf{E-VQA}} &
\multicolumn{3}{c}{\textbf{InfoSeek}} \\
\cmidrule (lr){4-5}\cmidrule(lr){6-8}
& & & \textbf{Single-Hop} & \textbf{All} & \textbf{Unseen-Q} & \textbf{Unseen-E} & \textbf{All} \\
\midrule
\rowcolor{black!8}
\multicolumn{8}{l}{\textit{Zero-shot MLLMs}} \\
BLIP-2\cite{blip2} &Flan-T5$_\mathrm{XL}$&-&12.6&12.4&12.7&12.3&12.5\\
InternVL3-8B\cite{internvl3} &InternVL3-8B&-&15.7&15.9&13.5&11.7&12.5\\
Qwen2.5-VL-7B\cite{qwen2.5vl} &Qwen2.5-VL-7B&-&16.3&16.3&16.8&15.8&16.3\\
Qwen3-VL-8B\cite{qwen3vl} &Qwen3-VL-8B&-&19.5&19.5&19.7&17.4&18.5\\
GPT-4V\cite{gpt4v} &GPT-4V&-&26.9&28.1&15.0&14.3&14.6\\
\midrule
\rowcolor{black!8}
\multicolumn{8}{l}{\textit{Retrieval-Augmented Models}} \\
DPR$_\mathrm{V+T}$$^{\dagger}$\cite{dpr} &Multi-passage BERT&CLIP ViT-B/32&29.1&-&-&-&12.4\\
EchoSight$^{\dagger}$ \cite{echosight}&Mistral-7B/LLaMA-3-8B&EVA-CLIP-8B&19.4&-&-&-&27.7\\
Wiki-LLaVA\cite{wikillava} &Vicuna-7B&CLIP ViT-L/14+Contriever&17.7&20.3&30.1&27.8&28.9\\
Prophet++$^{\ast}$ $^{\lozenge}$$^{\dagger}$ \cite{prophet}&Qwen3-VL-8B&EVA-CLIP-8B&22.5&20.0&13.2&11.6&12.3\\
NoteMR$^{\lozenge}$$^{\dagger}$ \cite{notemr}&Qwen3-VL-8B&EVA-CLIP-8B&25.6&23.6&28.7&29.8&29.2\\
ReflectiVA$^{\ast}$  \cite{reflectiva}&LLaMA-3.1-8B&EVA-CLIP-8B&28.0&29.2&40.4&39.8&40.1\\
MMKB-RAG \cite{mmkbrag}&Qwen2-7B&EVA-CLIP-8B&39.7&35.9&36.4&36.3&36.4\\
VLM-PRF$^{\ast}$  \cite{vlmprf}&InternVL3-8B&EVA-CLIP-8B&40.1&39.2&43.5&42.1&42.5\\
% \midrule
% \multicolumn{4}{l}{\textit{Our Proposed Method}} \\
\rowcolor{yellow!10}
\textbf{MaS-VQA(Ours)}$^{\dagger}$ &InternVL3-8B&EVA-CLIP-8B&40.8&40.1&42.9&42.6&42.8\\
\rowcolor{yellow!10}
\textbf{MaS-VQA(Ours)}$^{\dagger}$ &Qwen3-VL-8B&EVA-CLIP-8B&\textbf{42.2}&\textbf{41.3}&\textbf{43.7}&\textbf{43.9}&\textbf{43.8}\\

\bottomrule
\end{tabular}}

\label{tab:main_results}
\end{table*}

% \begin{table}[t]
% \caption{Performance on the OK-VQA. Best results are highlighted in bold.}
% \centering
% \small
% \setlength{\tabcolsep}{3pt}
% \begin{tabular}{l c c}
% \toprule
% \textbf{Method}&\textbf{Model}&\textbf{Score} \\
% \midrule
% ASB&LLaMA 2 (13B)&61.2 \\
% Prophet&GPT-3 (175B)&61.2 \\
% SKP&Vicuna-7B&63.3\\
% NoteMR&LLaVA-NeXT-8B&70.0\\
% MaS-VQA(Ours)&& \\
% \bottomrule
% \end{tabular}
% \label{tab:ablation}
% \end{table}

\begin{table}[t]
\caption{Ablation study on Encyclopedic-VQA (test) assessing the impact of attention masking, phrase selection, and implicit knowledge. Best results are highlighted in bold.}
\centering
\small
\setlength{\tabcolsep}{3pt}
\resizebox{\columnwidth}{!}{%
\begin{tabular}{c c c c c}
\toprule
\multicolumn{2}{c}{\textbf{Explicit Knowledge}} &
\multirow{2}{*}{\textbf{Implicit Knowledge}} &
\multicolumn{2}{c}{\textbf{E-VQA}} \\
\cmidrule (lr){1-2}
\cmidrule (lr){4-5}
\textbf{Attention Mask}&\textbf{Phrase Selection} & & \textbf{Single-Hop} & \textbf{All} \\
\midrule
\checkmark &&&39.8&36.8\\
&\checkmark&&38.5&36.6\\
\checkmark&\checkmark&&40.9&38.4\\
&&\checkmark&40&35.9\\
\checkmark&\checkmark&\checkmark&\textbf{42.2}&\textbf{41.3}\\
\bottomrule
\end{tabular}}
\label{tab:ablation}
\end{table}

\begin{table}[t]
\centering
\small
\setlength{\tabcolsep}{10pt}
\caption{Effect of retrieval breadth on Encyclopedic-VQA. MaS-VQA performance as the number of retrieved passages $k$ varies. The retriever and MLLM backbone are fixed; only $k$ is changed.}
\begin{tabular}{c cc}
\toprule
\multirow{2}{*}{\textbf{$k$}} & \multicolumn{2}{c}{\textbf{E-VQA}} \\
\cmidrule(lr){2-3}
 & \textbf{Single-Hop} & \textbf{All} \\
\midrule
1  & 35.1 & 31.6 \\
3  & 39.5 & 36.7 \\
5  & \textbf{42.2} & \textbf{41.3} \\
7 & 41.8 & 41.1 \\
\bottomrule
\end{tabular}
\label{tab:evqa_k_sweep}
\end{table}

\subsection{Experimental Setup}
\textbf{Dataset.} We conduct experiments on two knowledge-based visual question answering (KB-VQA) benchmark datasets: E-VQA and InfoSeek.
E-VQA contains 2.21 million visual question–answer pairs, each associated with up to five images sourced from iNaturalist 2021 and GLDv2. The dataset is split into approximately 1 million training samples, 13.6k validation samples, and 5.8k test samples. Questions are categorized into single-hop and two-hop. The accompanying knowledge base consists of around 2 million image-centric Wikipedia articles covering 16,700 fine-grained entities. Following standard practice, we evaluate on the official test set.
InfoSeek is a large-scale information retrieval–based VQA dataset with 1.3 million image–question pairs built upon approximately 11,000 visual entities from OVEN. It is divided into 934k training samples, 73k validation samples, and 348k test samples, where the validation set is further split into unseen-entity and unseen-question subsets to assess generalization. Following the official evaluation protocol, we report results on the complete validation set with no overlap in questions or entities with the training data.

\textbf{Baselines.} We categorize the baselines into two groups: zero-shot multimodal large language models (MLLMs) and retrieval-augmented models. The zero-shot MLLMs include BLIP-2\cite{blip2}, InternVL3-8B\cite{internvl3}, Qwen2.5VL-7B\cite{qwen2.5vl}, Qwen3-VL-8B\cite{qwen3vl} and GPT-4V\cite{gpt4v}. The retrieval-augmented models comprise DPR\cite{dpr}, EchoSight\cite{echosight}, Wiki-LLaVA\cite{wikillava}, Prophet++\cite{prophet}, NoteMR\cite{notemr}, ReflectiVA\cite{reflectiva}, MMKB-RAG\cite{mmkbrag}, and VLM-PRF\cite{vlmprf}.

\begin{figure*}[t]
  \centering
  \includegraphics[width=\textwidth]{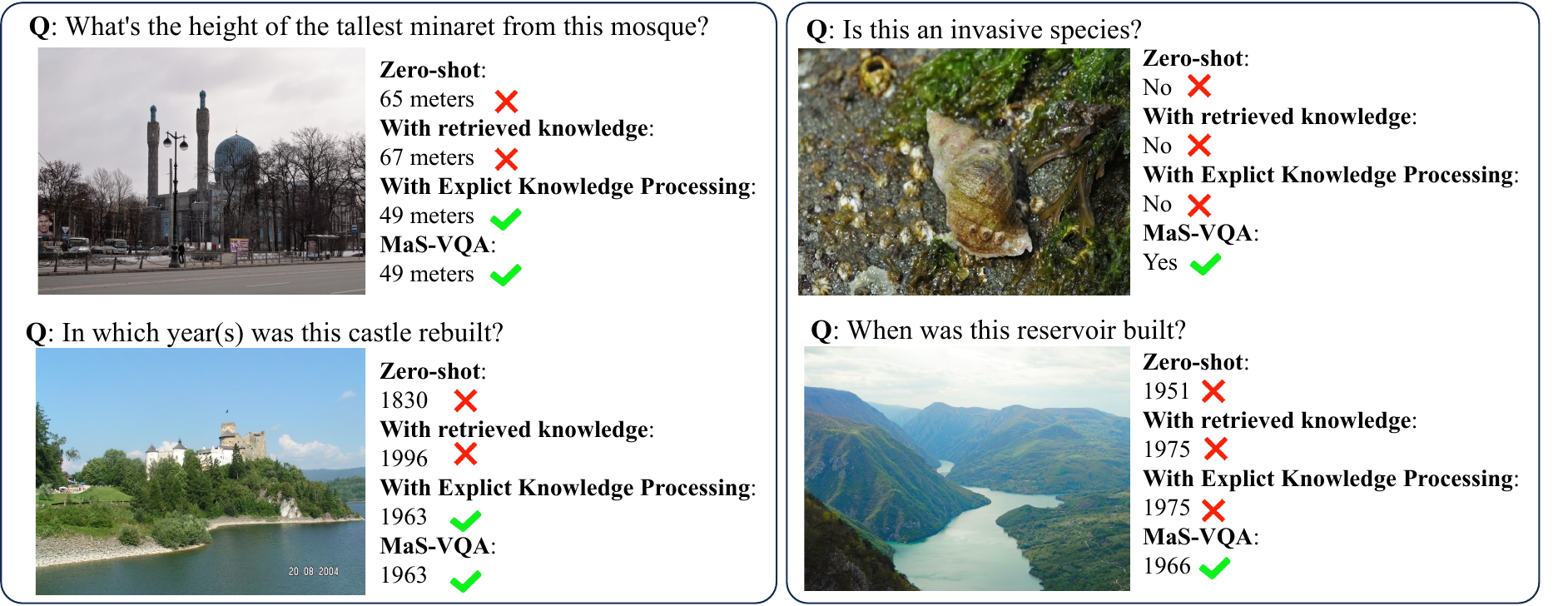}
  \caption{Qualitative case studies. Left: explicit external knowledge helps bridge missing factual gaps and corrects errors made without retrieval. Right: implicit knowledge complements retrieved evidence when the final decision requires commonsense/domain priors beyond the retrieved text.}
  \label{fig:casefigure_1}
\end{figure*}

\textbf{Evaluation Metrics.} Following the official evaluation protocols of each dataset, we adopt dataset-specific metrics to ensure fair comparison with prior work. For Encyclopedic-VQA, we use BERT Matching (BEM)~\cite{bem}, which judges the correctness of a predicted answer conditioned on both the question and the ground-truth answer, and is robust to semantically equivalent phrasings. For InfoSeek, we follow its standard split by question types: STRING and TIME questions are evaluated using VQA accuracy~\cite{vqascore} based on exact/normalized matching, while NUMERICAL questions are evaluated using Relaxed Accuracy~\cite{relaxedacc} with type-specific relaxation strategies to tolerate minor deviations in numeric values.

\textbf{Implementation Details.} We adopt InternVL3-8B\cite{internvl3} and Qwen3-VL-8B\cite{qwen3vl} as MLLM backbones. For consistency, we use the same backbone model for both the knowledge generator and the downstream general-purpose MLLM throughout the pipeline. In Section~3.1, we employ EchoSight~\cite{echosight} as the multimodal retriever and use the EchoSight knowledge base to query relevant knowledge, returning the top-$k$ retrieved entries with $k=5$. In Section~3.2, we use a pretrained BLIP model\cite{blip} for image--text encoding. All experiments are conducted on NVIDIA H20 GPUs with FlashAttention-2\cite{flashattention2} enabled. Unless otherwise specified, inference uses a temperature of 0.7, a maximum generation length of 512 tokens, and a batch size of 16 to accelerate decoding. More implementation details are provided in Appendix~\ref{app:experimentalsetup}.

\subsection{Main Results}
\label{sec:main_results}

Table~\ref{tab:main_results} summarizes the main results on Encyclopedic-VQA (test) and InfoSeek (validation), with the best numbers highlighted in bold. Overall, \textbf{MaS-VQA} delivers the strongest performance across both benchmarks and all reported splits, indicating the effectiveness of selection-driven explicit-implicit knowledge co-modeling.

\paragraph{Results across benchmarks.}
On Encyclopedic-VQA, MaS-VQA markedly outperforms zero-shot MLLMs that rely solely on parametric knowledge: with Qwen3-VL-8B, it improves from 19.5 (zero-shot) to \textbf{42.2} (Single-Hop) and \textbf{41.3} (All), demonstrating that external knowledge is critical for encyclopedic questions and that our selection mechanism can effectively distill useful evidence from noisy retrieval. MaS-VQA is also competitive with, or better than, strong retrieval-augmented baselines (e.g., MMKB-RAG and VLM-PRF), achieving the best overall accuracy on the full test set. 
On InfoSeek, MaS-VQA attains the best results on Unseen-Q, Unseen-E, and All; with the same backbone, it reaches \textbf{43.7}, \textbf{43.9}, and \textbf{43.8}, respectively, indicating strong generalization to novel questions and entities by leveraging compact, question-relevant evidence rather than brittle lexical overlap or noisy retrieval.

\paragraph{Effect of backbone models.}
MaS-VQA consistently improves performance with both InternVL3-8B and Qwen3-VL-8B, suggesting that the gains stem from our MaS-VQA framework rather than a specific backbone.

\begin{figure*}[t]
  \centering
  \includegraphics[width=0.9\textwidth]{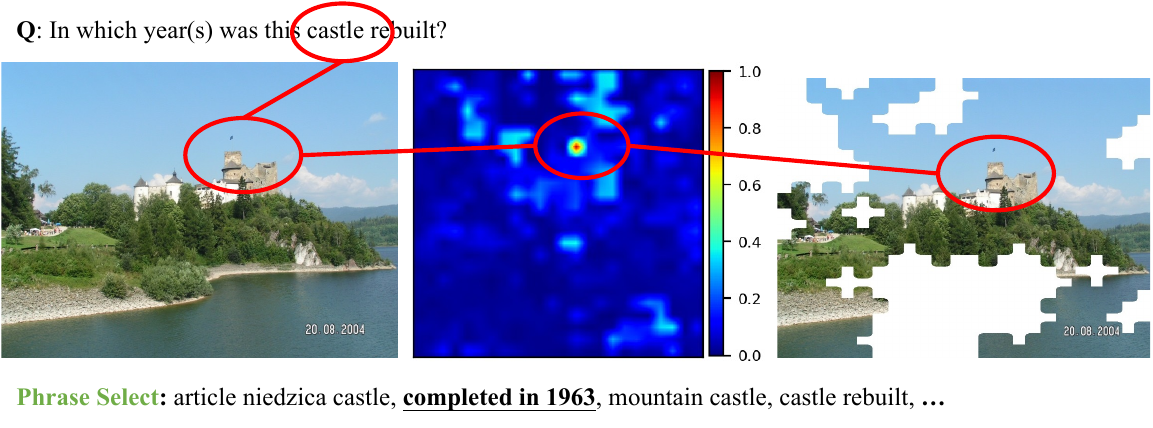}
  \captionsetup{skip=3pt}
  \caption{Implicit Knowledge Processing complements explicit evidence. Even with filtered explicit knowledge, some questions require additional commonsense/domain priors. Our implicit knowledge processing elicits such parametric knowledge conditioned on the selected evidence, enabling correct reasoning and answers.}
  \label{fig:casefigure_2}
\end{figure*}

\begin{figure*}[t]
  \centering
  \includegraphics[width=0.9\textwidth]{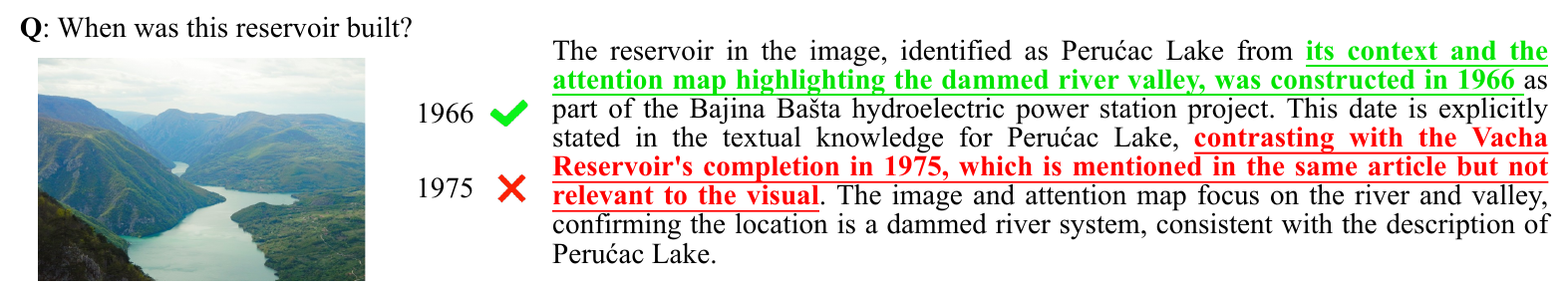}
  \caption{Implicit Knowledge Processing complements explicit evidence. Even with filtered explicit knowledge, some questions require additional commonsense/domain priors. Our implicit knowledge processing elicits such parametric knowledge conditioned on the selected evidence, enabling correct reasoning and answers.}
  \label{fig:casefigure_3}
\end{figure*}

\paragraph{Discussion on methods requiring additional training.}
In Table~\ref{tab:main_results}, $^{\ast}$ denotes methods that require additional training during the reasoning stage. While such approaches typically increase computation and may reduce fairness when compared to purely inference-time methods, MaS-VQA improves reasoning \emph{without} additional training at inference time by selecting and compressing retrieved knowledge and question-relevant visual cues to better elicit implicit (parametric) knowledge from a frozen backbone.

\subsection{Ablation Study}
Table~\ref{tab:ablation} studies the contribution of each component in MaS-VQA (Qwen3-VL-8B) on Encyclopedic-VQA (test). Under the explicit-knowledge setting, enabling the question-guided attention mask alone yields 39.8/36.8 (Single-Hop/All), while using phrase selection alone achieves 38.5/36.6. Combining the two explicit grounding modules further improves performance to 40.9/38.4, indicating that visual-region localization and question-conditioned phrase selection provide complementary signals for constructing higher-quality explicit evidence. When only implicit knowledge is used (i.e., without explicit grounding), the performance drops to 40.0/35.9, suggesting that parametric knowledge alone is less reliable and can be distracted by unfiltered or redundant inputs. Finally, integrating implicit knowledge with both explicit modules achieves the best results, reaching \textbf{42.2}/\textbf{41.3}, which confirms that MaS-VQA benefits from jointly modeling explicit retrieval-grounding and implicit knowledge distillation.

We further analyze the effect of retrieval breadth in Table~\ref{tab:evqa_k_sweep}. Increasing the number of retrieved passages from $k=1$ to $k=5$ consistently improves performance (35.1/31.6 $\rightarrow$ 42.2/41.3), showing the importance of having sufficient evidence coverage. However, further increasing $k$ to $7$ leads to a slight drop (41.8/41.1), indicating that overly broad retrieval may introduce additional noise that offsets the benefit of more passages. Based on this trade-off, we set $k=5$ in all experiments.

\subsection{Case Study}
\textbf{Effectiveness of Explicit Knowledge Processing.}
As shown in Figure~\ref{fig:casefigure_1} (left), directly appending retrieved passages often introduces redundant or weakly relevant evidence, which may distract the model with spurious associations and ultimately mislead answer selection.
With our \emph{Explicit Knowledge Processing}, MaS-VQA performs joint modality filtering: a knowledge-guided attention mask suppresses irrelevant image regions and encourages the model to attend to the question-critical visual cues, while question-conditioned phrase selection prunes noisy or off-topic textual fragments and keeps only the most informative spans.
Consequently, the final context becomes more compact, better aligned across vision and language, and easier for the MLLM to reason over. Figure~\ref{fig:casefigure_2} further visualizes this effect---the model concentrates on the correct visual region and retains the key phrase from retrieved knowledge---providing an intuitive explanation for the consistent gains over raw retrieval prompting.

\textbf{Effectiveness of Implicit Knowledge Processing.}
Figure~\ref{fig:casefigure_1} (right) presents challenging cases where explicit filtering alone remains insufficient, e.g., when answering requires additional commonsense/domain priors or implicit relational reasoning beyond the retrieved snippets.
In these cases, \emph{Implicit Knowledge Processing} further elicits and organizes parametric knowledge conditioned on the filtered evidence, enabling MaS-VQA to complete the missing reasoning step and correct the prediction.
A representative example is provided in Figure~\ref{fig:casefigure_3}, where implicit knowledge complements the cleaned explicit evidence to reach the correct answer.

%% file: content/5_conclusion.tex
We presented \textbf{MaS-VQA}, a selection-based framework for KB-VQA that addresses the noise and distraction introduced by raw retrieval prompting. By jointly filtering question-relevant visual regions and retrieved knowledge via a \textbf{Mask-and-Select} mechanism, MaS-VQA constructs low-noise explicit evidence aligned with the question. Building on this filtered context, MaS-VQA further elicits complementary \emph{implicit} knowledge conditioned on the selected evidence, enabling more reliable reasoning. Experiments on E-VQA and InfoSeek show consistent improvements over strong baselines across different MLLM backbones, while ablations and case studies verify our method’s effectiveness.

\section*{Impact Statement}
This work studies knowledge-based visual question answering (KB-VQA) by improving how external retrieved knowledge and implicit knowledge are selected and combined. The proposed selection mechanism can reduce the influence of noisy or irrelevant evidence, which may improve robustness and interpretability in knowledge-intensive multimodal applications such as educational assistants, accessibility tools, and information-seeking systems.

Potential negative impacts include the possibility of amplifying biases present in the underlying knowledge base or the MLLM, and producing plausible but incorrect answers when retrieval is incomplete or when implicit knowledge is miselicited. Our method does not guarantee factual correctness; therefore, it should not be used as the sole basis for high-stakes decisions (e.g., medical, legal, or safety-critical settings). We encourage future work on stronger evidence verification, uncertainty estimation, and bias evaluation/mitigation for KB-VQA systems.

%% file: content/6_appendix.tex
\section{Test Dataset Statistics}
Table~\ref{tab:dataset_stats} reports the dataset statistics for our evaluation splits. 

\textbf{Templated.} Templated questions are manually created by domain experts based on predefined super-categories, resulting in controlled and standardized question patterns.
\textbf{Automatic.} Automatic questions are generated by feeding Wikipedia articles into a question generation model, producing more diverse question formulations.
\textbf{Multi-answer.} For multi-answer questions, the authors first use automatically generated questions to produce an initial candidate answer list, and then ask human annotators to complete and verify all possible correct answers.
\textbf{Two-Hop.} Two-hop questions are constructed using a \emph{bridge entity}: the answer to the first single-hop question is used as the subject of a second single-hop question, forming a two-step reasoning chain.
\begin{table}[H]
\centering
\small
\setlength{\tabcolsep}{8pt}
\resizebox{\columnwidth}{!}{
\begin{tabular}{lllrr}
\toprule
\textbf{Split} & \multicolumn{2}{c}{\textbf{Question Type}} & \textbf{\#Questions} & \textbf{Total} \\
\midrule
\multirow{4}{*}{E-VQA Test}
  & \multirow{3}{*}{Single-Hop} & Templated     & 1000 & \multirow{4}{*}{5750} \\  && Automatic     & 2750 &  \\
  &     & Multi-answer  & 1000 &  \\
\cmidrule(lr){2-4}
  & Two-Hop & --  & 1000 &  \\
\midrule
InfoSeek Val & --   & --     & 71335 & 71335 \\
\bottomrule
\end{tabular}}
\caption{Dataset statistics of the Encyclopedic-VQA test set and the InfoSeek validation set.}
\label{tab:dataset_stats}
\end{table}

\section{Prompts}
\label{app:prompts}

\paragraph{Implicit Knowledge Synthesis.}
Given the explicit evidence (textual knowledge and extracted keywords) and the attention-guided visual signals (white regions are masked), the model produces a concise paragraph (2--5 sentences) that summarizes grounded insights useful for answering the question later. We explicitly instruct the model \textbf{not} to output the final answer in this stage.

\begin{tcolorbox}[  breakable,
  enhanced,
  colback=gray!10, colframe=black, arc=5pt, boxrule=1pt]
Synthesize implicit knowledge by integrating the four inputs below. Do not answer the question. Produce one concise paragraph (2–5 sentences) capturing the key grounded insights that would be useful for answering the question later. 

1. Evidence: Each Keyword paragraph is extracted from the Knowledge paragraph above it.

    Textual Knowledge: Textual information relevant to answering the question. You need to identify information that helps you answer the question.
    
    Keywords: Keywords selected based on the textual knowledge to assist your answer.
    
2. Original Image: An image relevant to the question.

3. Attention Map: Visualization of key areas based on the original image. White areas are masked; you only need to focus on the non-white areas.
\vspace{\baselineskip}

Evidence: \{\textit{evidence}\}

Original Image: \textless image\textgreater

Attention Map: \textless image\textgreater

Question: \{\textit{question}\}

Implicit Knowledge:
\end{tcolorbox}

\paragraph{Final Answer Prediction.}
The model answers the question using (i) the selected explicit evidence, (ii) the synthesized implicit knowledge, and (iii) the visual inputs. The attention map is treated as supplementary grounding information to reduce distractions from irrelevant regions. The output is constrained to a single word or short phrase to match the evaluation format.

\begin{tcolorbox}[  breakable,
  enhanced,
  colback=gray!10, colframe=black, arc=5pt, boxrule=1pt]
Your task is to perform visual question answering based on the given text and image. I will provide you with the following information:

1. Evidence: Each Keyword paragraph is extracted from the Knowledge paragraph above it.

    Textual Knowledge: Textual information relevant to answering the question. You need to identify information that helps you answer the question.
    
    Keywords: Keywords selected based on the textual knowledge to assist your answer.
    
2. Original Image: An image relevant to the question.

3. Attention Map: Visualization of key areas based on the original image. White areas are masked; you only need to focus on the non-white areas.

Please use the evidence and original image as global information, and the attention map as supplementary information to answer the question.
\vspace{\baselineskip}

Evidence: \{\textit{evidence}\}

Implicit Knowledge: \{\textit{imknowledge}\}

Original Image: \textless image\textgreater

Attention Map: \textless image\textgreater

Question: \{\textit{question}\}

Answer the question using a single word or phrase.

Answer:
\end{tcolorbox}

\section{Experimental Setup}
\label{app:experimentalsetup}
\textbf{Implementation Details.}
We adopt InternVL3-8B~\cite{internvl3} and Qwen3VL-8B~\cite{qwen3vl} as the frozen MLLM backbones. For consistency, the same backbone is used for both implicit knowledge generation ($f_{\mathrm{imp}}$) and final answer inference ($g_{\theta}$) throughout the pipeline, and we do not update MLLM parameters in any stage.

\textbf{Retrieval and explicit knowledge construction.}
For explicit textual evidence retrieval ($f_{\mathrm{ret}}$), we use EchoSight~\cite{echosight} as the multimodal retriever to query the external knowledge base $\mathcal{K}$, and keep the top-$k$ passages with $k=5$, forming $T=\{t_i\}_{i=1}^{k}$. Retrieved passages are concatenated with the question into a single sequence
$X=\texttt{[CLS]}\;T\;\texttt{[SEP]}\;Q\;\texttt{[SEP]}$.
When the input exceeds the maximal text length of the encoder ($\texttt{max\_txt\_len}=512$), we truncate $X$ accordingly. We enable fast tokenization and store character-level offset mappings to support phrase span recovery.

\textbf{Grounding encoder and attention-based signals.}
To derive explicit grounding signals, we employ a pretrained BLIP image--text matching (ITM) encoder~\cite{blip} (\texttt{blip\_image\_text\_matching\_large}) to implement $f_{\mathrm{mask}}$ and $f_{\mathrm{key}}$. For the image-side attention mask $M$, we extract cross-attention weights from transformer block $b=7$ and compute gradient-weighted relevance maps (attention $\times$ ReLU(gradient)) with respect to the positive ITM logit, followed by head averaging and per-token min--max normalization. We adopt adaptive token reweighting with softmax temperature $\tau=1.0$, including intra-group token normalization within the knowledge span and the question span, as well as group-level balancing between knowledge and question tokens. Patch-level binarization is performed via token-wise percentile thresholding with percentile $\rho=90$. Token masks are combined using a logical \textsc{OR} operation across tokens (\texttt{combine\_strategy=any}) to obtain the final $g\times g$ patch mask, which is then upsampled to the image resolution and visualized by suppressing non-salient regions with a white background.

For the text-side keyword set $\mathbf{k}$, we compute gradient-modulated text self-attention interactions (question-to-knowledge) at the same transformer block ($b=7$) with \texttt{output\_attentions=True}. We score each knowledge token by averaging interaction scores over all question tokens and select the top-$m$ tokens with $m=30$. Selected token indices are mapped back to character spans in the retrieved passages using offset mappings and merged into readable phrases with a span merge gap of $3$ character. We keep the first $10$ merged phrases as the final keyword set $\mathbf{k}$.

\textbf{Inference configuration and hardware.}
All experiments are conducted on NVIDIA H20 GPUs with FlashAttention-2~\cite{flashattention2} enabled. Unless otherwise specified, we use a temperature of $0.7$, a maximum generation length of $512$ tokens, and a batch size of $16$ for decoding acceleration.

\section{Additional Case Study}
\begin{figure*}[t]
  \centering
  \includegraphics[width=\textwidth,height=0.42\textheight,keepaspectratio]{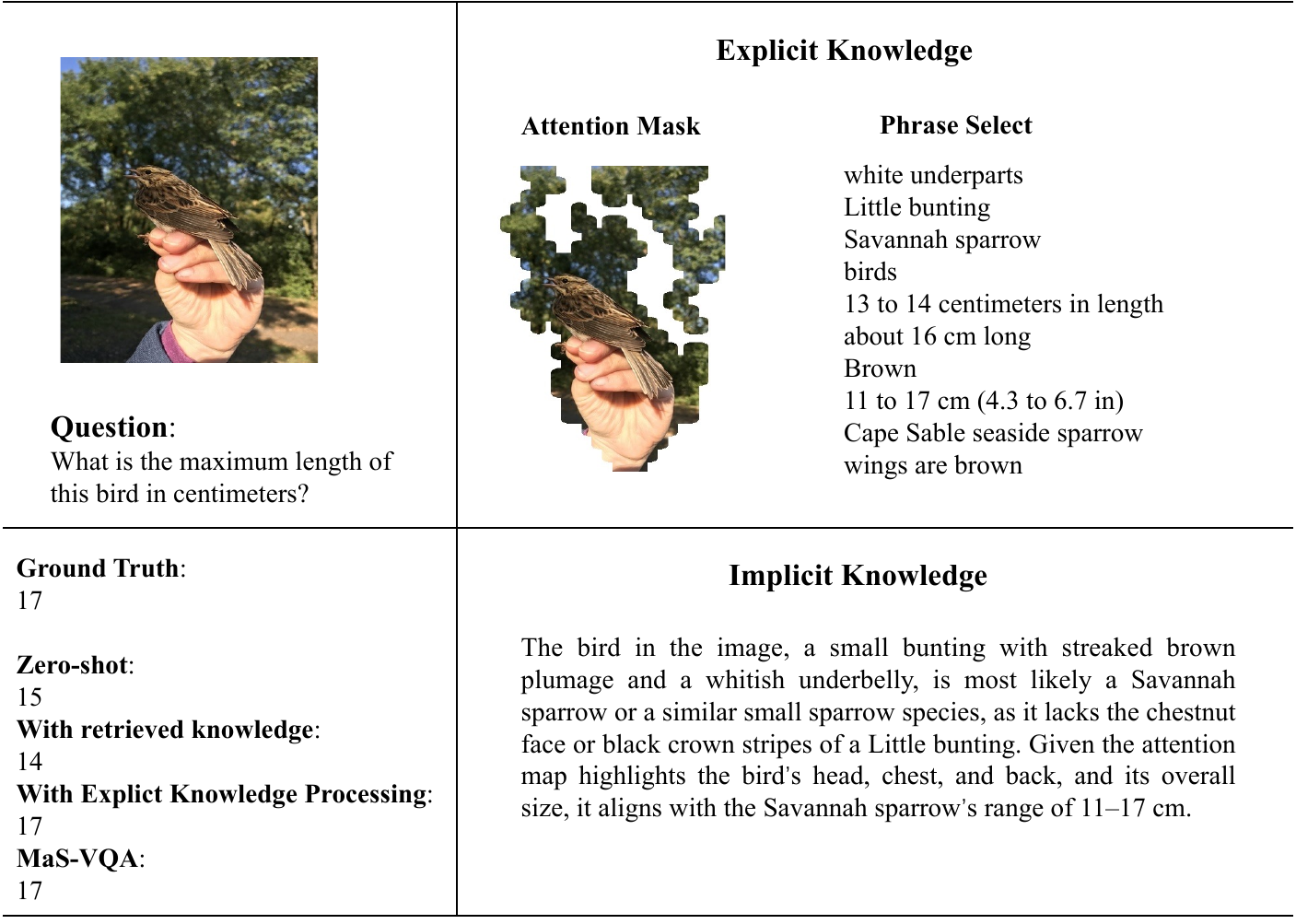}\par\vspace{6pt}
  \includegraphics[width=\textwidth,height=0.42\textheight,keepaspectratio]{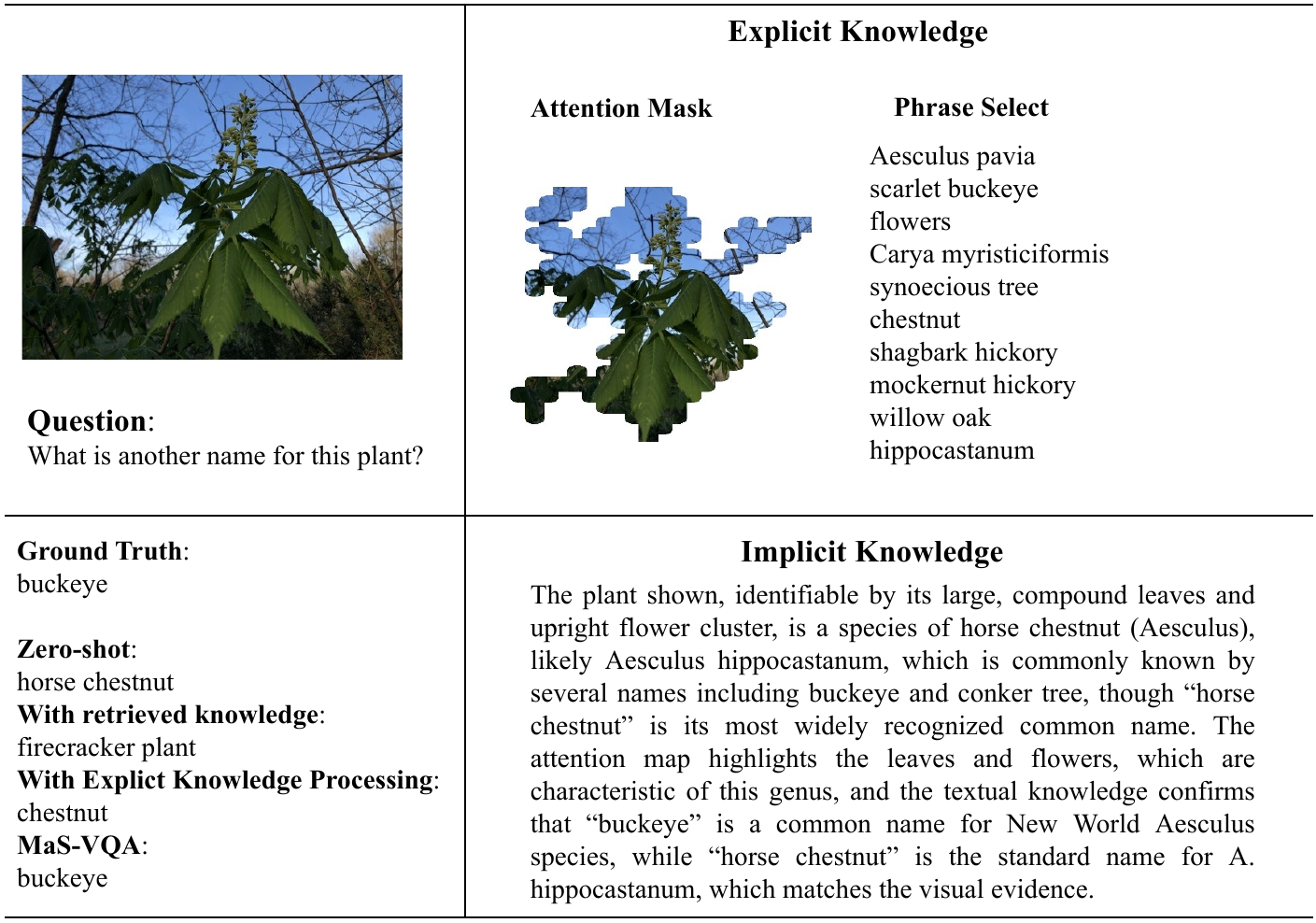}
  \caption{Case studies of MaS-VQA.}
  \label{fig:ap12}
\end{figure*}

\begin{figure*}[t]
  \centering
  \includegraphics[width=\textwidth,height=0.42\textheight,keepaspectratio]{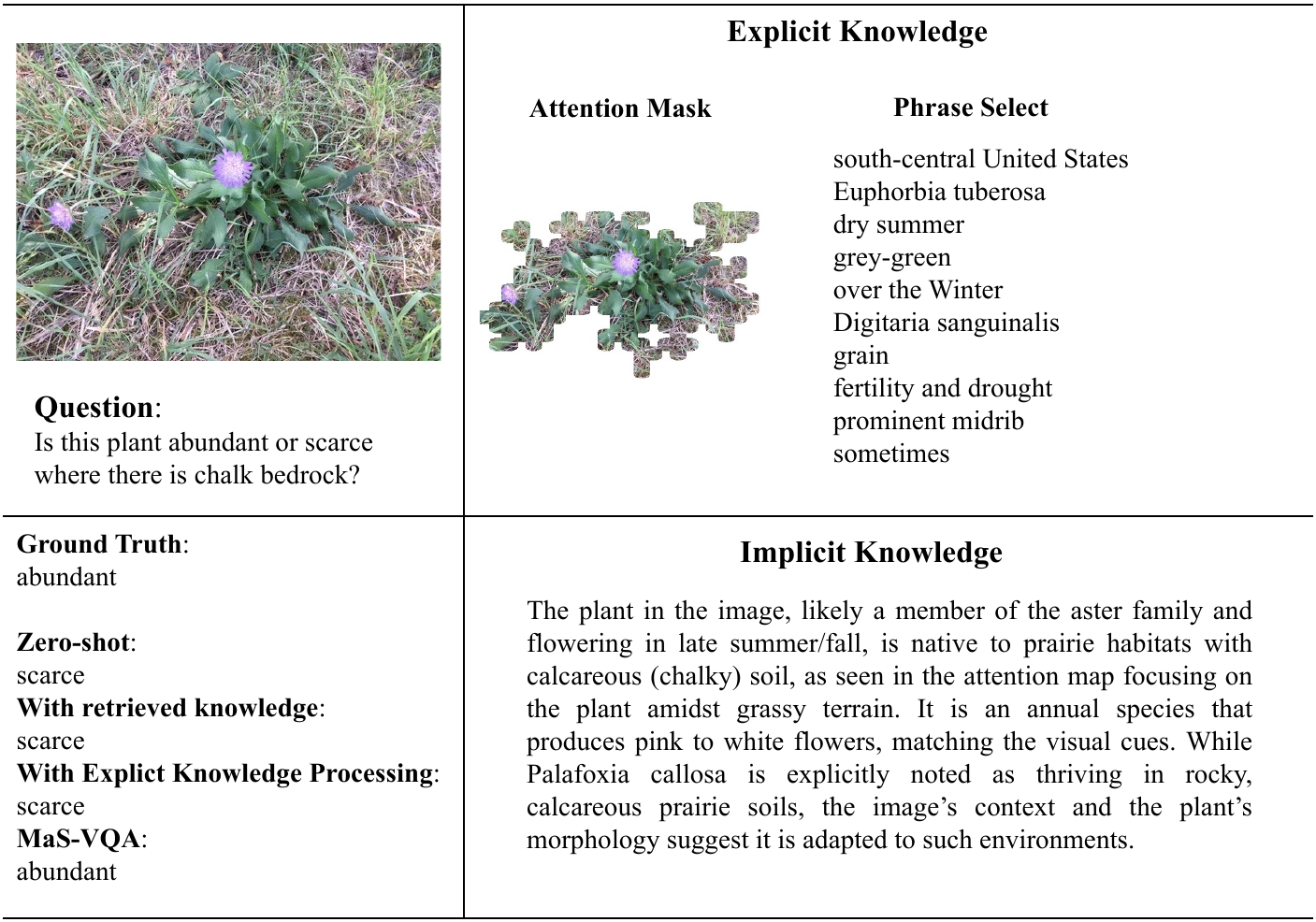}\par\vspace{6pt}
  \includegraphics[width=\textwidth,height=0.42\textheight,keepaspectratio]{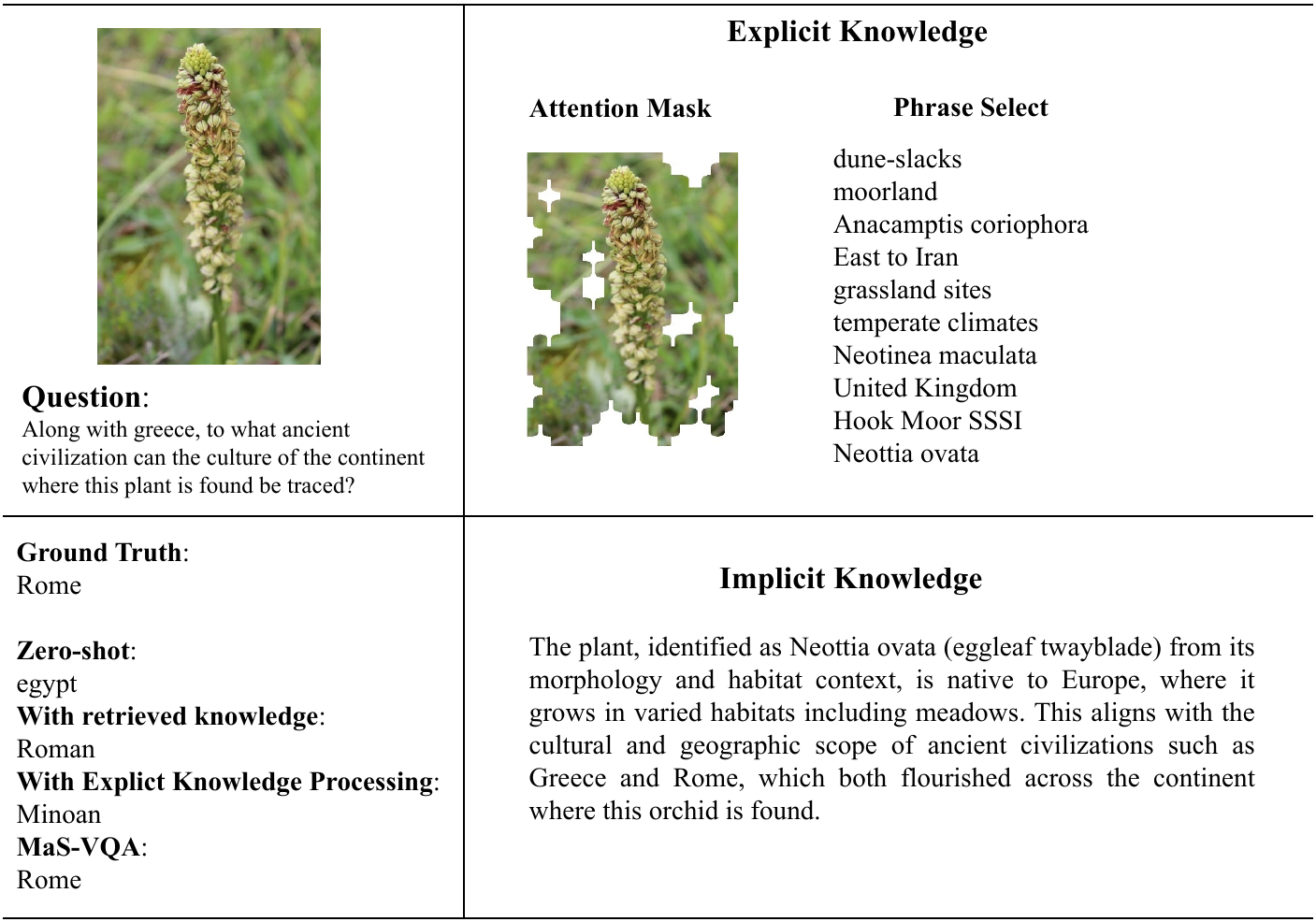}
  \caption{Case studies of MaS-VQA.}
  \label{fig:ap34}
\end{figure*}

To provide a deeper understanding of MaS-VQA, we include additional and more detailed qualitative case studies in this section. These examples further illustrate the behavior of the model. As shown in Figs.~\ref{fig:ap12} and~\ref{fig:ap34}, MaS-VQA effectively grounds retrieved knowledge to question-relevant image regions and leverages the generated implicit knowledge to support robust multi-step reasoning.

